\documentclass[preprint,journal]{vgtc}       

\ifpdf
  \pdfoutput=1\relax                   
  \usepackage{graphicx}                
  \DeclareGraphicsExtensions{.pdf,.png,.jpg,.jpeg} 
\else
  \usepackage{graphicx}                
  \DeclareGraphicsExtensions{.eps}     
\fi%

\graphicspath{{figures/}{pictures/}{images/}{./}} 

\usepackage{microtype}                 
\PassOptionsToPackage{warn}{textcomp}  
\usepackage{textcomp}                  
\usepackage{mathptmx}                  
\usepackage{times}                     
\usepackage{cite}                      
\usepackage{tabu}                      
\usepackage{booktabs}                  
\usepackage{amssymb}
\usepackage{subfigure}
\usepackage{url}

\usepackage{breakurl}
\usepackage[breaklinks]{hyperref}
\usepackage{tcolorbox}
\definecolor{tmcolor}{rgb}{0.7,0,0}

\preprinttext{Preprint}

\onlineid{1220}

\vgtccategory{Research}
\vgtcpapertype{analytics/decisions}

\title{FuNNscope: Visual microscope for interactively exploring the loss landscape of fully connected neural networks}

\author{Aleksandar Doknic and Torsten Möller}
\authorfooter{
\item
 Aleksandar Doknic and Torsten Möller are with University of Vienna. E-mail: \{aleksandar.doknic\,$|$\,torsten.moeller\}@univie.ac.at.
}

\shortauthortitle{Biv \MakeLowercase{\textit{et al.}}: Global Illumination for Fun and Profit}

\abstract{Despite their effective use in various fields, many aspects of neural networks are poorly understood. One important way to investigate the characteristics of neural networks is to explore the loss landscape. However, most models produce a high-dimensional non-convex landscape which is difficult to visualize. We discuss and extend existing visualization methods based on 1D- and 2D slicing with a novel method that approximates the actual loss landscape geometry by using charts with interpretable axes. Based on the assumption that observations on small neural networks can generalize to more complex systems and provide us with helpful insights, we focus on small models in the range of a few dozen weights, which enables computationally cheap experiments and the use of an interactive dashboard. We observe symmetries around the zero vector, the influence of different layers on the global landscape, the different weight sensitivities around a minimizer, and how gradient descent navigates high-loss obstacles.
The user study resulted in an average SUS (System Usability Scale) score with suggestions for improvement and opened up a number of possible application scenarios, such as autoencoders and ensemble networks.%
} 

\keywords{Loss landscape, explainable models, neural network}

\teaser{
  \centering
  \includegraphics[width=\linewidth]{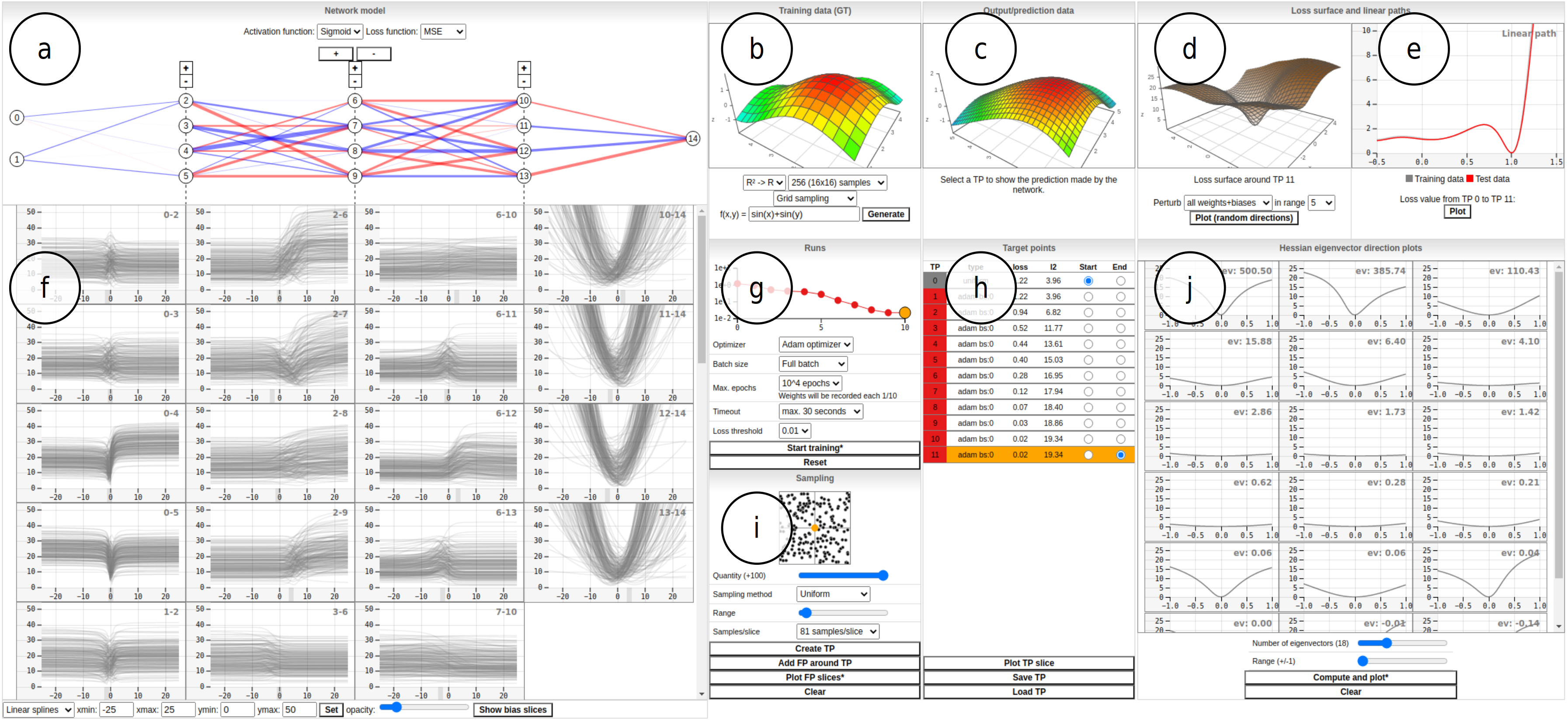}
  \caption{FuNNscope GUI: The network is trained to approximate the function $f(x,y) = \sin(x)+\sin(y)$. (a) shows a representation of the trained network model, (b) training data, (c) network prediction, (d) dimensionality-reduced loss landscape, (e) interpolated loss (MSE) between initial weight vector and final weight vector, (f) loss surface from different "perspectives" computed for each weight, (g) training loss over epochs, (h) loss and weight norm at different stages of training, (i) low-dimensional representation of sampling points used to produce (f), (j) loss curves in Hessian eigenvector directions.}
  \label{fig:teaser}
}

\vgtcinsertpkg

\begin{document}


\firstsection{Introduction}

\maketitle

Training a deep neural network (DNN) \cite{DBLP:books/daglib/0040158} is, in theory, a highly complex problem due to the large number of weight parameters and intrinsic non-convexity of the loss surface, but in practice, DNN can often be trained effectively with simple algorithms like stochastic gradient descent (SGD). By investigating the loss landscape (e.g. \cite{DBLP:journals/corr/abs-2012-06898,DBLP:conf/aistats/ChoromanskaHMAL15,DBLP:journals/corr/ImTB16,DBLP:conf/nips/Li0TSG18,DBLP:conf/aaai/FortS19,DBLP:journals/corr/abs-1804-10200,DBLP:journals/corr/abs-2104-11044,DBLP:journals/corr/abs-2106-16004,DBLP:journals/corr/abs-2006-05900}) and its relation to network architecture, training parameters, and generalizability, researchers try to get a better understanding of the behaviour of DNNs.
\newpage
Different visualization methods (e.g. \cite{DBLP:journals/corr/abs-2012-06898,DBLP:conf/nips/Li0TSG18}) have been used for empirical experiments, but they only show a fraction of the true complexity, and even though 2D slices \cite{DBLP:conf/nips/Li0TSG18} of a high-dimensional loss surface look familiar and intuitive, they only show a thin slice of the full parameter space.

We propose a novel loss landscape visualization method based on axis-parallel slicing \cite{DBLP:journals/cgf/Torsney-WeirSM17} that approximates a geometrically accurate representation of the loss landscape for small network models by systematically sampling the area around interesting weight vectors, in particular the minimizers found by training algorithms.

In this work, we create a proof of concept to demonstrate that axis-aligned slices are an effective approach to understanding local and global behavior of loss landscapes, consciously leaving the scale-up problem to millions of weights for future research. We take inspiration from the success of the interactive visualization tool Tensorflow Playground \cite{playground} which allows users to interact directly with small models and uses visualizations to provide an intuitive understanding of DNNs. By focusing our approach on small networks, we created interactive visualizations with human-in-the-loop workflows and a relatively complete view of the loss landscape for all parameter dimensions. With the integration of different loss visualization methods in a single dashboard, we hope to provide a deeper and more intuitive understanding of the loss landscape of DNNs.

Our contributions include: 1) Surveying and discussing common loss landscape visualization methods and their context within machine learning (ML) research, 2) developing and extending the slicing approach \cite{DBLP:journals/cgf/Torsney-WeirSM17} for neural loss surfaces and putting the different visualization methods in context with each other, 3) designing and implementing a dashboard prototype that enables the user to run experiments on small networks and investigate the loss landscape with different methods and from different perspectives, 4) producing three use cases where we apply the prototype to investigate the loss landscape (i) from a global perspective, (ii) around a minimizer, and (iii) along a gradient descent trajectory, and drawing conclusions from the visualizations, 5) testing and discussing the prototype with ML practitioners to validate the usability, gain feedback of the different visualizations, and receive input for possible future applications, and 6) reflecting on the results of our work by discussing the advantages of axis-parallel slicing for our use cases, addressing issues for larger models, and potential directions for future research.

The paper is structured as follows: \autoref{ch:background} gives a brief description of the loss function. \autoref{ch:relatedwork} discusses insights from prior publications, state-of-the-art methods, and challenges of visualizing optimization surfaces in the context of neural loss landscapes. The axis-parallel slicing approach is discussed in \autoref{ch:slicingapproach}. In \autoref{ch:observationtargets} we describe important aspects of the loss landscape that are of interest to researchers and the reasons behind it.
Goals and tasks are defined in \autoref{ch:goalsandtasks}, and the visualization design is discussed in \autoref{ch:visualization}. Three use cases are explored in detail in \autoref{ch:usecases}. In the user evaluation in \autoref{ch:evaluation} we ask ML practitioners to explore the landscape with our prototype and comment on their user experience, which is also quantified by the standardized System Usability Scale (SUS) questionnaire. Finally, we discuss our findings and lessons learned in \autoref{ch:discussion} and summarize our results and contributions in \autoref{ch:conclusion}.

\section{Background} \label{ch:background}

We define the loss as the error of the DNN output with respect to a training data set. The averaged loss value at point $\mathbf{\theta}$ (network weights vector) for $N$ input data points can be defined as $L(\mathbf{\theta}) = \frac{1}{N} \sum_{i=1}^N  l(x_i,y_i;\mathbf{\theta})$ with input data $\mathbf{x}$ and respective ground truth labels $\mathbf{y}$. For regression tasks the loss function $l(x_i,y_i;\mathbf{\theta})$ is commonly defined as the mean square error (MSE) or mean absolute error (MAE). Loss landscape plots can be created by systematically sampling $L(\mathbf{\theta})$ at different weight vectors $\theta$, e.g. along a 1D trajectory or a 2D plane.
The input data generally remains the same during the sampling process, although some training algorithms (SGD and mini-batch gradient descent) use a subset of the full training data for each iteration. 

\section{Related work} \label{ch:relatedwork}

One method of visualizing the loss landscape and its minima was described by Goodfellow et al. \cite{DBLP:journals/corr/GoodfellowV14}. \textit{Linear interpolation experiments} plot the loss value starting from the initial parameters $\theta_i$ along the linearly interpolated SGD trajectory with the final training parameters $\theta_f$. The goal is to see whether the cross-section is well-behaved. They argue that plotting a simple loss curve during training
does not show whether the loss landscape is bumpy or if there is stochastic noise due to the gradient estimation in SGD. They also argue that it does not show if a loss remains constant for a long time due to being stuck in a flat region, oscillating around a local minimum, or tracing around an obstacle. Therefore they propose plotting the loss $L(\theta)$ on a linear path with $\theta = (1-\alpha)\theta_0+\alpha\theta_1$ where $\theta_0$ represents the starting parameters and $\theta_1$ represents the final parameters of the linear path plot. In experiments with the MNIST dataset and a fully connected feed forward network (maxout, sigmoid and ReLU) they show that, if $\theta_0$ is the initial weights and and $\theta_1$ is the trained parameters, then the loss function is approximately convex along the linear path. This means that a line search would be highly effective in training such networks if we knew the correct direction. However, it has been shown that it is possible to construct networks that violate this property~\cite{DBLP:journals/corr/abs-2104-11044}. 

\begin{figure}[t]
    \centering
    \includegraphics[width=0.5\linewidth]{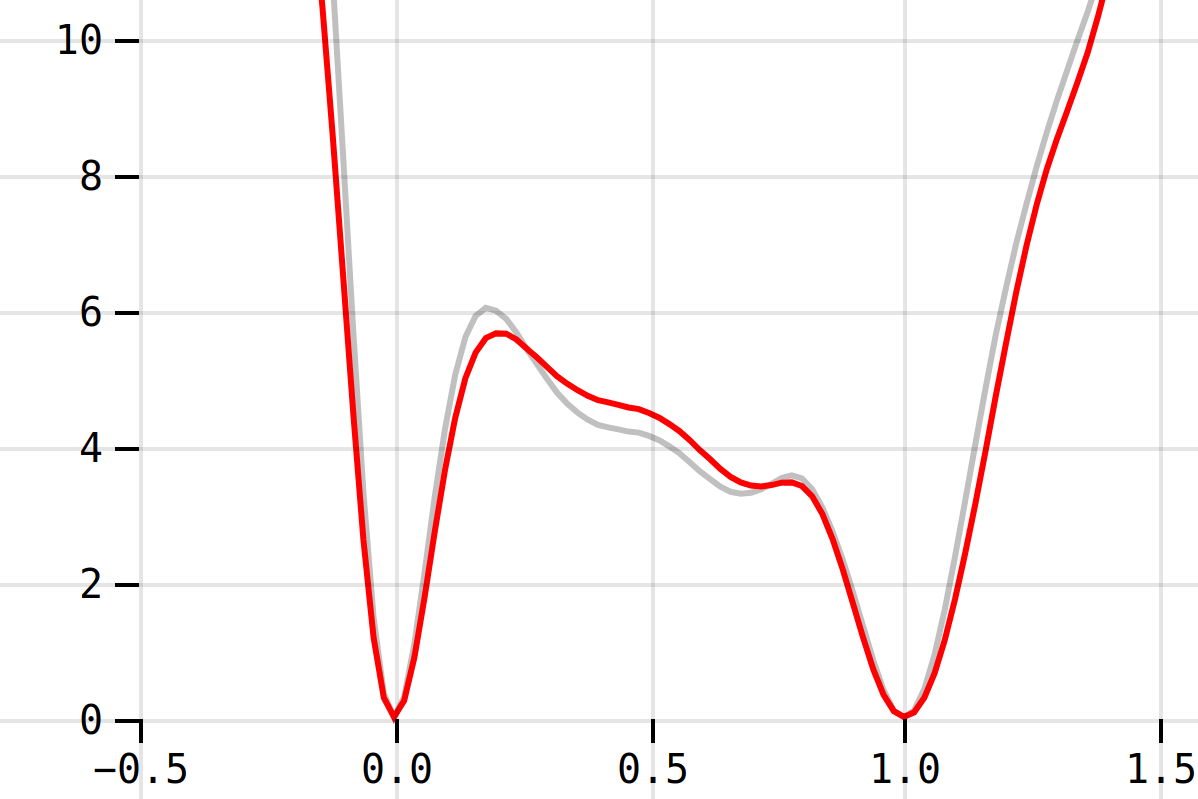}
    \caption{Linear interpolation shows the loss value (vertical axis) along a linear path (horizontal axis) between two minimizers. The horizontal axis represents the interpolation coefficient. The grey curve represents training loss, while the red curve represents test loss. Note that both curves overlap at the minima, which indicates that the minima generalize better than the barrier between them.}
    \label{fig:lin2d}
\end{figure}

\autoref{fig:lin2d} shows how two minima can be compared along the 1D trajectory (by setting $\theta_0$ and $\theta_1$ to the respective values), which was used to compare the qualitative characteristics of minima found by different optimization algorithms \cite{DBLP:conf/nips/Li0TSG18}.
Li et al. \cite{DBLP:conf/nips/Li0TSG18} also show that weight decay and batch size influence the flatness and steepness characteristics of minimizers. They propose the method of \textit{filter-wise normalization} to make minima comparable.

\begin{figure}[t]
    \centering
    \includegraphics[width=0.4\linewidth]{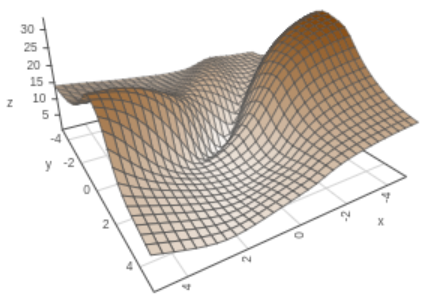}
    \caption{Loss value on 2D plane spanned by random vectors around minimizer exposes a landscape of apparent minima and saddle points.}
    \label{fig:plot3d}
\end{figure}

\begin{figure}[t]
    \centering
    \includegraphics[width=1.0\linewidth]{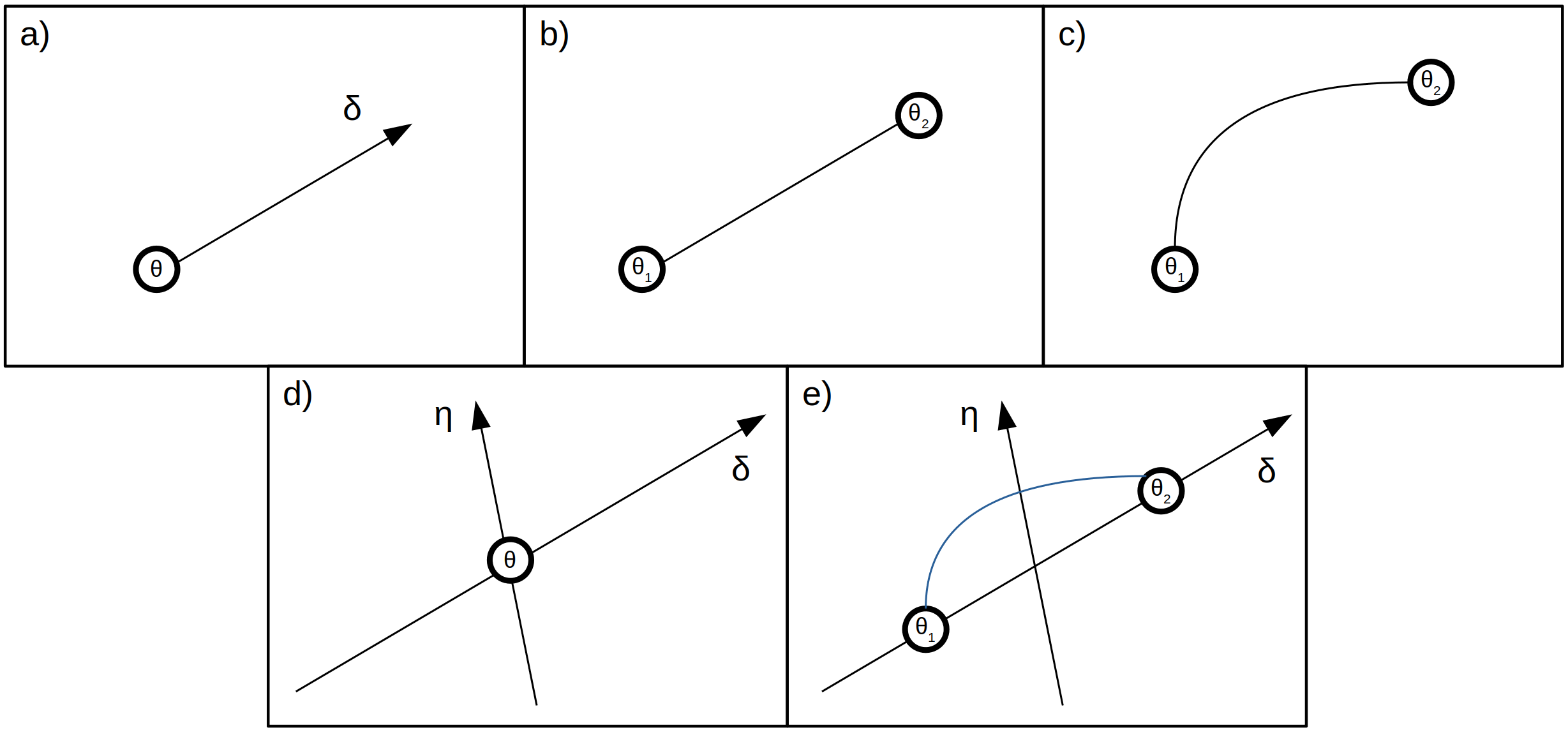}
    \caption{Five methods for visualizing a low-dimensional representation of the loss surface of a DNN: a) Multiples of vector $\delta$ are added to starting weight vector $\theta$ incrementally, plotting a linear path into the given direction. b) The loss along the linear interpolation path between vectors $\theta_0$ and $\theta_1$ is plotted. c) The loss value is plotted along a non-linear path (e.g. SGD trajectory) between $\theta_0$ and $\theta_1$. d) The loss surface is plotted on a plane spanned by vector $\delta$ and $\eta$, usually with a minimizer $\theta$ in the center. e) A plane intersecting $\theta_0$ and $\theta_1$ is spanned by $\delta$ and $\eta$ which are chosen in such a way, that a path of low loss (e.g. minimum energy path \cite{DBLP:conf/icml/DraxlerVSH18}) is visible. Note that this list is not complete.}
    \label{fig:dir}
\end{figure}

Another method \cite{DBLP:conf/nips/Li0TSG18} is based on creating a 2D plane spanned by two random vectors centered around the minimum, see \autoref{fig:plot3d}. The loss value $L(\theta)$ is computed for $\theta = \theta_0+\alpha\delta+\beta\eta$. $\alpha$ and $\beta$ are scalars that describe the size of the plane with random direction vectors $\delta$ and $\eta$ around the weight vector $\theta_0$. This is represented by \autoref{fig:dir} (d). Analyzing the surface with filter normalization leads to the observation that wider models and those with skip connections show less chaotic behavior, flatter minima, and wide regions of apparent convexity. Chaotic landscapes result in worse training and test errors, while more convex landscapes generalize significantly better. Li et al. \cite{DBLP:conf/nips/Li0TSG18} show that 2D plots with random directions actually represent convexity by computing the eigenvalue ratio $|\lambda_{min}/\lambda_{max}|$ of the Hessian across a random 2D plane and observing that negative eigenvalues remain extremely small in convex-looking surfaces, which justifies the random direction visualization approach.

It is also possible to choose the plane in such a way that it shows two minima. One hypothesis about loss landscapes states that solutions of over-parametrized networks lie on a connected (sub)manifold~\cite{DBLP:journals/corr/abs-1804-10200,DBLP:journals/corr/abs-2003-00307}. This was visualized by Draxler et al. \cite{DBLP:conf/icml/DraxlerVSH18}, which use heuristic approximations to find the \textit{minimum energy path}. These patterns are not visible on the linear interpolation approach, which often shows barriers of high loss between two minima.

Adaptions of the methods discussed in this chapter have also been used to plot a non-linear interpolation of the loss surface between three and four weight vectors (e.g. initial weights or minimizer) \cite{DBLP:journals/corr/ImTB16}.

LossPlot \cite{DBLP:journals/corr/abs-2111-15133} is a recent loss landscape visualization tool that enables the user to do experiments and provides a comparison of up to six loss landscape visualizations that are produced for different network settings. However, it is currently limited to the 2D slice method \cite{DBLP:conf/nips/Li0TSG18}.

While there is a tremendous amount of work in the ML community focused on loss landscapes, all of the previous work is focused on a single local 1D or 2D slice. We are the first ones that suggest a more comprehensive global characterization of the loss landscape, supported by an interactive, exploratory visual analysis.

\section{The slicing approach} \label{ch:slicingapproach}

\begin{figure}[tp]
    \centering
    \subfigure[Himmeblau's function]{\includegraphics[width=0.16\textwidth]{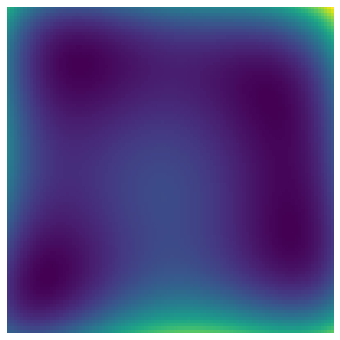}}
    \subfigure[Random $x_1$ slices]{\includegraphics[width=0.16\textwidth]{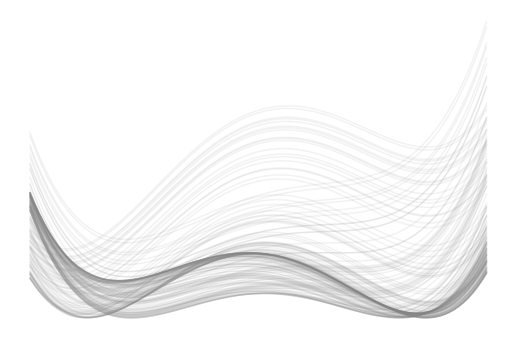}} 
    \subfigure[Random $x_2$ slices]{\includegraphics[width=0.16\textwidth]{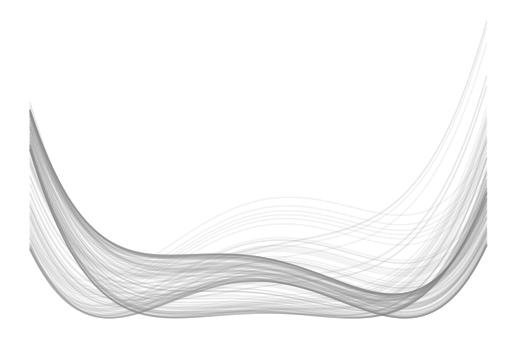}}
    \caption{Himmelblau's function \cite{himmelblau}: The slice visualizations (b) and (c) provide an alternative to the heatmap visualization (a).}
    \label{fig:himmelblau}
\end{figure}

The idea of the slicing approach \cite{DBLP:journals/cgf/Torsney-WeirSM17} is to visualize a $D$-dimensional function as an ensemble of $D$ 1-dimensional slices where the horizontal axis describes the parameter value while the vertical axis describes the function value. For example, we can see that Himmelblau's~\cite{himmelblau} function can be visualized as a heatmap in \autoref{fig:himmelblau} (a), but also as overlapping axis-parallel slices in \autoref{fig:himmelblau} (b) and (c). The practical advantage of this method is that it can be applied to functions with more than three dimensions. It can replace dimensionality reduction methods that produce less intuitive axes, and low-dimensional slices that conceal geometric information.

In the case of $f: \mathbb{R}^2 \rightarrow \mathbb{R}$ such as \autoref{fig:himmelblau} we intuitively understand the slices as being \textit{in front} or \textit{behind} each other with respect to the other axis. In higher dimensional space, we have to acknowledge that each slice in one dimension depends on the combination of all remaining parameters. The path from one overlapping slice to another is expected to be on a diagonal, which means non-axis-parallel parameter changes are implicitly included. 

\begin{figure}[t]
    \centering
    \subfigure[Single focus point on Ackley's function \cite{ackley1987model}]{\includegraphics[width=0.16\textwidth]{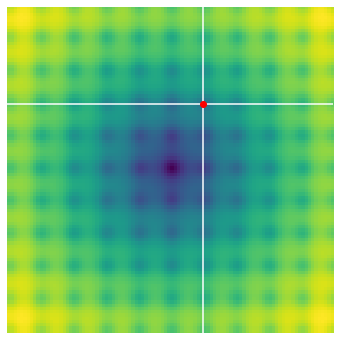}} 
    \subfigure[Slice at $x_1=1$]{\includegraphics[width=0.16\textwidth]{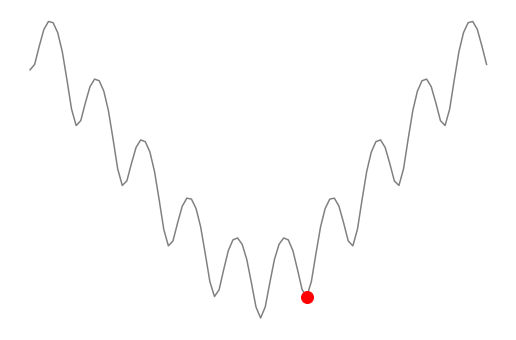}}
    \subfigure[Slice at $x_2=2$]{\includegraphics[width=0.16\textwidth]{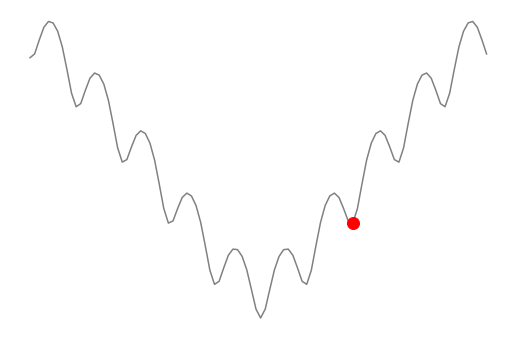}}\\
    \subfigure[Multiple focus points]{\includegraphics[width=0.16\textwidth]{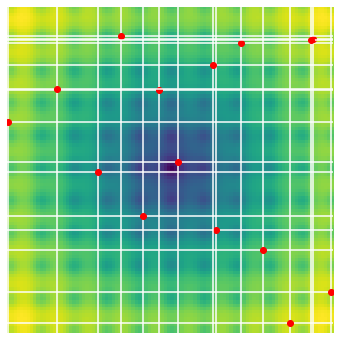}} 
    \subfigure[Random $x_1$ slices]{\includegraphics[width=0.16\textwidth]{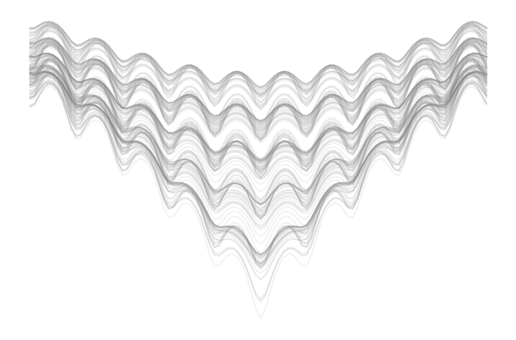}} 
    \subfigure[Random $x_2$ slices]{\includegraphics[width=0.16\textwidth]{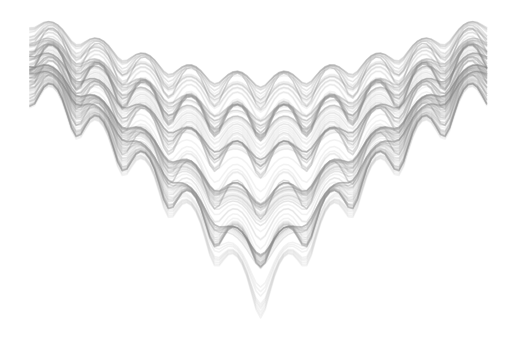}}
    \caption{From a single point in Ackley's function \cite{ackley1987model} two 1D loss slices in (b) and (c) can be plotted. Sampling focus points such as in (d) reveals even more information about the function structure, leading to visualizations similar to (e) and (f).}
    \label{fig:ackley}
\end{figure}

\textit{Focus points} \cite{DBLP:journals/cgf/Torsney-WeirSM17} are vectors in the input parameter space. In the case of DNNs, focus points are weight vectors. Each weight vector represents a network state. \textit{Slicing} means that one parameter is iteratively changed while all other parameters remain fixed. \autoref{fig:ackley} (a) shows the slice intersection at the focus point. Along each of these slices, the loss value can be computed and visualized, which is shown in \autoref{fig:ackley} (b) and (c). Adding multiple focus points, such as in \autoref{fig:ackley} (d) reveals more details about the function landscape, which is shown in \autoref{fig:ackley} (e) and (f). While each focus point represents only one point on the surface, the slices cover the whole parameter space within the given range.

\section{Observation targets} \label{ch:observationtargets}
Networks are usually initialized with weights sampled from a distribution before they are trained. The training algorithm follows a trajectory through the high-dimensional space until it converges to an acceptable loss value. This section describes important points along the training trajectory that could be investigated in detail.

\subsection{Initial weights}
Initialization choice significantly influences the training success in DNNs. The \textit{Goldilock's Zone} \cite{DBLP:conf/aaai/FortS19} describes an initialization space with preferable characteristics and a high level of convexity where good optimization results can be expected, while initial parameter vectors outside the Goldilocks zone increase training difficulty. Using the linear interpolation approach, Vlaar et al. \cite{DBLP:journals/corr/abs-2106-16004} find that different layers have varying sensitivity to the choice of initialization parameters.

\subsection{Saddle points}
Im and Tao \cite{DBLP:journals/corr/ImTB16} hypothesize that optimization algorithms arrive at different minima due to making characteristic choices at saddle points. Saddle points are also one of the possible reasons why second-order algorithms without modifications struggle compared to first-order algorithms \cite{DBLP:books/daglib/0040158}. In fact, it appears that the ratio of saddle points to local minima increases exponentially with the dimensionality \cite{DBLP:conf/nips/DauphinPGCGB14}.

\subsection{Minimizers}
Algorithms \cite{DBLP:conf/nips/HochreiterS94,DBLP:journals/neco/HochreiterS97a} were designed to find flat minima (i.e. the loss value is insensitive to small parameter changes) based on the assumption that they provide a better generalization than steep minima. One possible explanation is provided by the minimum description length theory \cite{DBLP:journals/automatica/Rissanen78} which motivates the idea that flat minimizers generalize better because they can be specified with lower precision \cite{DBLP:conf/iclr/KeskarMNST17}. Additionally, SGD training algorithms using small batch sizes consistently converge to flatter minima compared to using larger batch sizes, for both training data and validation data, due to inherent gradient noise \cite{DBLP:conf/iclr/KeskarMNST17}. However, Dinh et al. \cite{DBLP:conf/icml/DinhPBB17} show that it is possible to build equivalent models with arbitrarily sharper minima that generalize well. Another interesting point of research is the behavior of optimization algorithms that converge to said minima.


\section{Goals and tasks} \label{ch:goalsandtasks}

Based on the related work literature on loss landscapes we defined visualization goals and tasks that extend existing approaches with new points of view and allow users to casually set up experiments and explore the loss landscape. The idea is that small models can be computed faster and analyzed more thoroughly than large networks, which supports a trial-and-error approach for the development of research methods, experiments, and hypothesis validation.


\subsection{Goals}

\paragraph{G1-Geometric}
The user should be provided with visualizations that facilitate a deeper and \textit{global} understanding of the loss landscape. Previous techniques \cite{DBLP:journals/corr/GoodfellowV14,DBLP:conf/nips/Li0TSG18,DBLP:conf/iclr/KeskarMNST17} are based on observing the loss on single 1D or 2D slices of the parameter space. Our goal is to provide a \textit{global} perspective in terms of (a) sampling across the whole parameter range for each visualization and (b) providing different perspectives from each parameter by explicitly encoding the parameter values.

\paragraph{G2-Methodological consistency} The user should be able to view and study the same area with standard slicing techniques, in particular linear path experiments and plane slicing. The reasons are that users might be more familiar with these visualizations and find it easier to draw the connection to the other visualizations, and that methods could complement each other, i.e., one method could show what other visualization methods conceal. 

\paragraph{G3-Educational} The user should be able to design small networks and generate data for simple experiments. In particular, the network architecture, activation function, and loss function, as well as the training and test data, should be changeable. The network and the training data should be visualized in an intuitive way that helps the user understand the associated loss landscape. 

\paragraph{G4-Information} The user should be provided with complementary information, e.g., a list of different focus points with respective norm value, loss value, and other relevant information. The target user group (ML researchers and practitioners) should be familiar with the provided information.


\subsection{Tasks}

\paragraph{T1-Finding interesting points} The first task is to find interesting points that can be investigated. The most interesting point will be a minimizer, but it could also be a maximizer, a saddle point, random weights, or the zero vector (which is usually a bad choice for initializing networks). The prototype should offer simple methods for finding such interesting points, e.g., by sampling, choosing manually, or optimizing.

\paragraph{T2-Random plane experiments} Users can select a weight vector and plot the local area on a plane spanned by two vectors. The user can view the loss surface of the planes around different weight vectors. This visualization addresses G2-Methodological consistency.

\paragraph{T3-Slicing experiments} The user can create a slicing experiment by selecting a point and generating a selected number of focus points in the neighborhood with a chosen algorithm. Alternatively, the user can try to generate a global view by slicing a wide parameter range around the zero vector. 

\paragraph{T4-Linear path experiments} Users can compare minima by selecting two weight vectors and plotting the loss value on a linear path between them. The user can connect initial weights with trained weights, two initial weights, two minimizers, etc. The plot reveals, for example, relative sharpness and flatness, local convexity, loss barriers, and smoothness.



\subsection{Terminology}

Focus points are explained in \autoref{ch:slicingapproach}.
 Additionally, we define the \textit{target point} (TP) as the single weight vector that we are interested in investigating (i.e. the weight vector of the observation target). Of course, we could set a single focus point at the target point, but it would only show what happens if single parameters are changed. By adding multiple focus points and slicing the surrounding areas the cross-terms are implicitly included.

Note that literature sometimes refers to 2D surface plots as \textit{projections} \cite{DBLP:conf/nips/Li0TSG18,DBLP:journals/corr/abs-2104-11044,DBLP:journals/corr/ImTB16,DBLP:journals/corr/abs-2111-15133}. However, we believe this to be a misnomer. On the one hand, it is not a mathematical projection. Further, these visualizations do not "project" data points on a plane but rather represent the loss values that lie \textit{exactly} on a specific plane in the high-dimensional parameter space, similar to an MRI density image which shows the densities on a single plane in the human brain. Hence, we consider them 2D \textit{slices}. 

\section{FuNNscope}  \label{ch:visualization}
FuNNscope (\textit{fu}lly connected \textit{n}eural \textit{n}etwork micro\textit{scope})
is a web-based dashboard that was iteratively developed with the help of data scientists and visualization experts. The dashboard components have two purposes: the first is to create an experimental setup where a loss landscape can be explored, the second is to visualize the loss landscape itself in different ways.

\subsection{Experimental setup}
Loss landscape geometry depends on hyperparameters such as network architecture, activation function, and loss function, but also on the training data. 
The user is given the option to set up an experiment and create target points and focus points with the help of different menus and visualizations.
The menus and visualizations in this section are designed to meet the criteria of G3-Educational.
This approach was directly inspired by Tensorflow Playground \cite{playground}.

\paragraph{Network model}

The network model (\autoref{fig:teaser} (a)) represents enumerated neurons and weight connections. Positive weights are represented in red, negative weights in blue, and the line width encodes the magnitude. Neurons and layers can be added or removed by clicking on the respective buttons.

\paragraph{Training and prediction data}

The user can enter a 2D function to generate training data. The input parameters $x, y \in \mathbb{R}$ are sampled in the range $[0,5]$ and are used to create the ground truth from the user-specified function $f(x,y)$, see \autoref{fig:teaser} (b).
Test data is uniformly sampled in the same range. The target function and the network predictions are visualized in \autoref{fig:teaser} (b) and (c) on a 32x32 grid. This menu addresses the requirements of G4-Information.

\paragraph{Training (Runs)}

The user can select the number of epochs for a training run, including a timeout and a loss threshold as stopping conditions. If the training algorithm runs through all epochs, ten target points are created. Each target point represents network parameters, starting from the initial weights up to the, hopefully converged, final state. The training-generated target points are shown on the loss-epoch line chart (\autoref{fig:teaser} (g)).

\paragraph{Target points}

\autoref{fig:teaser} (h) shows how target points are represented in a list. Target points are weight vectors that have been obtained by random sampling (e.g. initial weights) or by training the network. They can also be stored and loaded from the disk. Hovering over a target point visually shows the weights in the network model. Clicking a target point sets the network to this state and shows the network output (\autoref{fig:teaser} (c)). After the network state is defined by a target point, the user can begin the loss landscape analysis on this point in parameter space.

\subsection{Loss landscape analysis}

FuNNscope offers, in total, four different perspectives on the loss landscape: \textit{Slice charts} are designed to represent the loss landscapes and the geometric structures for each parameter dimension, \textit{random vector plots} show the loss value on a 2D slice, \textit{linear interpolation plots} show the loss value that occurs when interpolating between two parameters, and the \textit{EV direction plots} show the loss values in the direction of the Hessian eigenvectors (EV).

\paragraph{Slice charts and sampling plot}

The slice charts (\autoref{fig:teaser} (f) and \autoref{fig:slices}) show a cross-section of the loss landscape along each axis dimension of the DNN, which is required by visualization goal G1-Geometric. A target point needs to be chosen, either in the training or target point view, as the center of the sampling space. Then the user can choose the number of focus points to create, the sampling algorithm, and the maximum sampling range relative to the target point in the center. The original Sliceplorer~\cite{DBLP:journals/cgf/Torsney-WeirSM17} uses Sobol sampling because it works on-the-fly and does not have to be recomputed when more focus points are added. However, we decided to let the user pick which algorithm to use due to the fact that different loss landscapes might have different requirements.

\begin{figure}[t]
    \centering
    \subfigure[Uniform]{\includegraphics[width=0.13\textwidth]{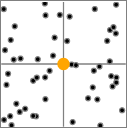}} 
    \subfigure[Sobol]{\includegraphics[width=0.13\textwidth]{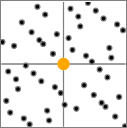}}
    \subfigure[Mixed]{\includegraphics[width=0.13\textwidth]{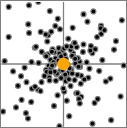}}
 \caption{Random focus points (grey) are sampled around the target point (orange). The user can zoom in and out and add focus points within different sampling ranges.}
    \label{fig:samplingplots}
\end{figure}

\autoref{fig:samplingplots} shows different sampling configurations. The sampling plot assists the choice of parameter settings and encodes the first two weight values for each focus point. It is, for example, possible to reduce the range and add additional focus points closer to the minimum, which is shown in \autoref{fig:samplingplots} (c). The user can also see differences between some sampling algorithms as shown in \autoref{fig:samplingplots} (a) and (b). 
The selected sampling range is also visualized as a dark region on the horizontal axis on all slice charts.

\paragraph{Random 2D slice plot}

The 3D plot is generated by calculating the loss value on a 2D plane spanned by random vectors. The random vectors are defined in such a way that they intersect the target point at the origin. For details see \autoref{ch:relatedwork}.

\paragraph{Linear interpolation plot}

Linear interpolation plots show the loss value between two target points. The loss curves for training data (grey) and test data (red) overlap when the network generalizes well. For details see \autoref{ch:relatedwork}.

\paragraph{EV direction plots}

Li et al.~\cite{DBLP:conf/nips/Li0TSG18} use heatmaps of the eigenvalue ratios to show that the 2D surface plots with random directions reveal convexity. FuNNscope also allows the user to compute the eigenvalues of the selected target point to study its convexity using Hessian-eigenthings~\cite{hessian-eigenthings}, and to compare the results to the other views. The slices are created by adding multiples of the approximated EVs of the Hessian matrix to the selected target point. In other words, they represent 1D slices along EV directions (instead of axis directions). The horizontal axis represents the direction of the EV. In our experiments, the EV plot associated with the largest eigenvalue reveals a strong curvature, which agrees with mathematical intuition.

\subsection{Design choices}

\begin{figure}[t]
    \centering
    \subfigure[Maximum opacity.]{\includegraphics[width=0.24\textwidth]{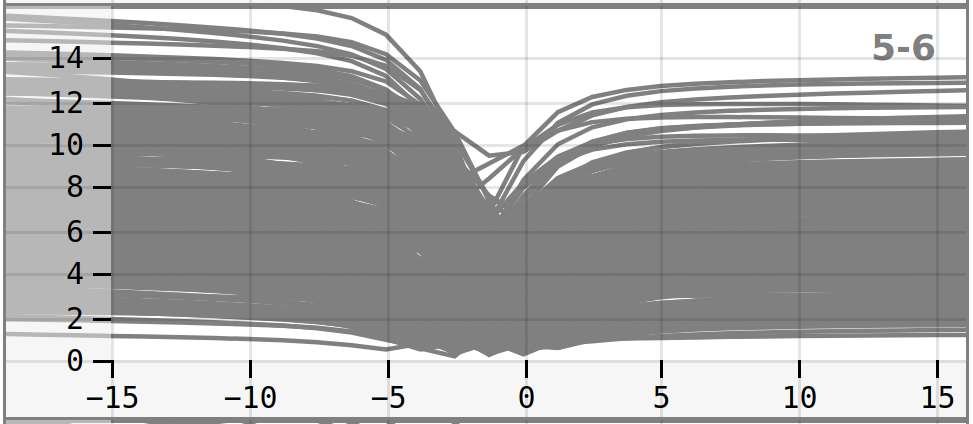}} 
    \subfigure[Reduced opacity.]{\includegraphics[width=0.24\textwidth]{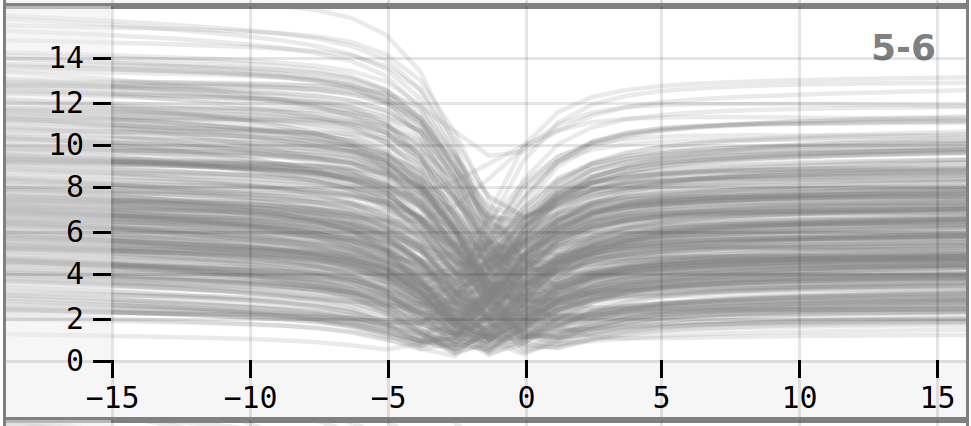}}\\
    \subfigure[Natural splines.]{\includegraphics[width=0.24\textwidth]{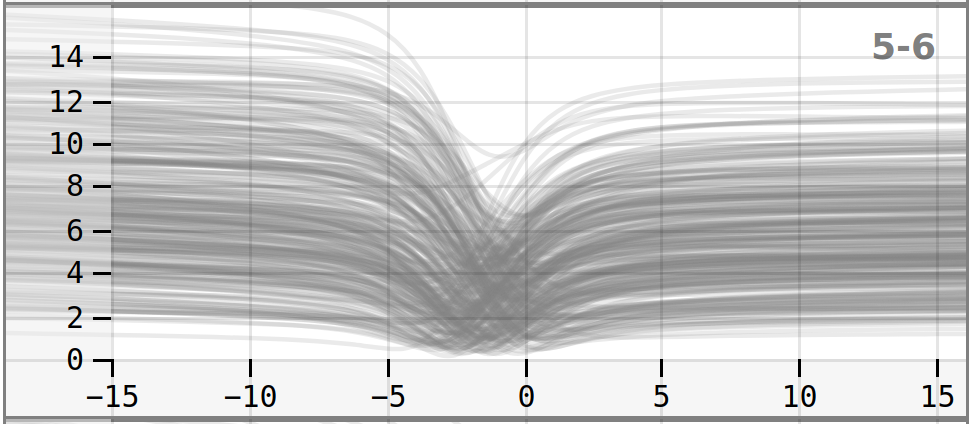}}
    \subfigure[Zoom on the minimizer (orange point).]{\includegraphics[width=0.24\textwidth]{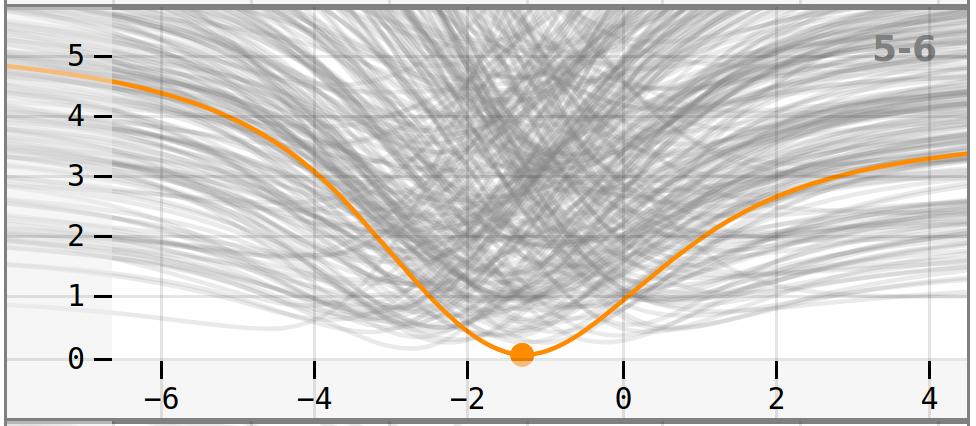}} 
    \caption{The slice chart for the weight between neuron 5 and 6 shows a distinct low loss region in the area where the minimizer is located.}
    \label{fig:slices}
\end{figure}

One significant advantage of the axis-parallel slice charts is that it shows an ensemble of 1D loss slices across the high-dimensional loss surface instead of a single 1D (interpolation methods) or 2D slice (as for the 2D slice plots). This deeper and more complete view that is shown in the slice charts requires a higher degree of visual flexibility compared to other visualizations in the dashboard. However, one gains a more global and robust understanding of the underlying high-dimensional loss function. During the prototyping process, we received formative feedback by ML practitioners and noticed visualization aspects that needed to be addressed.

The first visualization problem is that the amount of slices on the slice charts differs between experiments.
The reason is that the number of focus points required for a representative visualization depends on the complexity of the loss surface and the sampling range. Additionally, the slice density at different parts of the visualization can differ. Both aspects make it difficult to address overlaps equally for all experiments. One idea to resolve the overlapping issue was to blur out the single slices and melt them into a single visualization. However, too much information is lost, and both axes become harder to interpret. Another approach is to use a zoom function to separate the crowded slices. Zoom works well for inspecting parts of the visualizations (local exploration) but ignores the overview (global exploration). Finally, we decided, in addition to adding a zoom function, to let the user choose the opacity, which is shown in \autoref{fig:slices} (a) and (b). This allows the user to adapt the slice charts to different slice densities.

Another issue is that some slices exhibit high loss, and some stay close to zero. In order to address this problem, the user can change the axis scales. All slice charts share the settings. We avoided using different axis scales for different parameters because this would distort the perception of the loss surface for different dimensions, but it is possible to manually zoom into each slice chart.

Finally, the quality of the slice data can also differ. Fewer data points per slice lead to visible corners in areas of strong curvature. The user can increase the slice resolution in exchange for computation time in the sampling menu, but it might be preferable to use a fast method for a first analysis. The issue of low-resolution slices cannot be resolved directly because a complex loss surface function requires a sufficiently high slice resolution, but the option to switch from linear splines to natural splines allows the user to see a significantly smoother representation, especially in cases where few data points are available. \autoref{fig:slices} (c) shows how the corners that are visible around 0 on (a) and (b) become smooth. However, this approach only makes a guess of the loss value between points and should be taken with a grain of salt.

\subsection{Novelty}

The integrated view puts standard techniques such as random 2D slice plots and linear interpolation plots into a new context. In particular, the slice charts dismantle the deeply interwoven loss landscape information that is contained in the random 2D slice plots and split them into dimensional components. Additionally, the EV plots provide additional information about the loss landscape geometry.

\subsection{Implementation}

The dashboard is implemented as a JavaScript application with D3 \cite{d3js} and vis-graph3d \cite{vis-graph3d} for the visualizations. The backend is based on pytorch and hessian-eigenthings\cite{hessian-eigenthings} to compute the Hessian eigenvalues and eigenvectors. Linear interpolation experiments and 2D landscape were implemented as described in \autoref{ch:relatedwork} (without normalization filters).

\section{Use cases} \label{ch:usecases}

\begin{figure}[htp]
    \centering
    \includegraphics[width=0.5\textwidth]{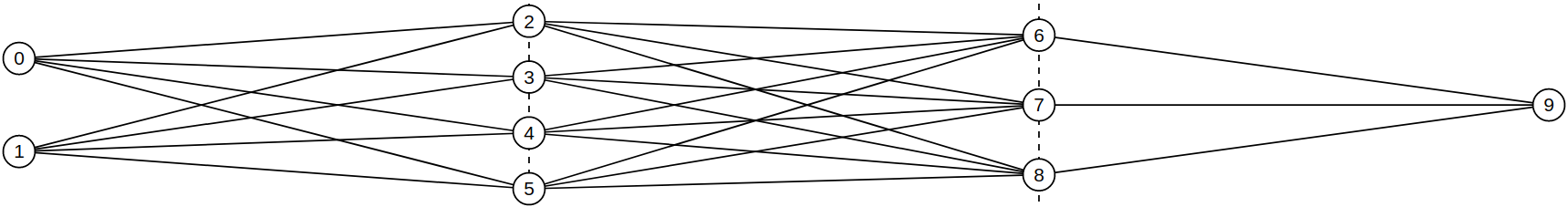}
    \caption{Sigmoid network for the use cases.}
    \label{fig:smallnetwork}
\end{figure}

For the following three use cases, we study the loss landscape of the Sigmoid network in \autoref{fig:smallnetwork}. It has two inputs and one output and is trained for a regression task. The two hidden layers consist of 4 and 3 neurons, resulting in 31 weights, 8 of which are bias weights. The training data consists of 256 random samples of $f(x,y) = \sin(x)+\sin(y)$ where the target variable is the function output at point $x, y \in [0,5]$. MSE is used as the loss.


\subsection{Preparation}

\begin{figure}[h]
    \centering
    \includegraphics[width=0.5\linewidth]{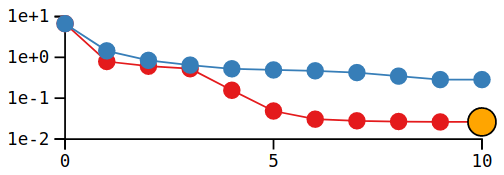}
    \caption{Target points are created during each training run. In this case, we see two training runs that created ten target points each. The last target point in the first (red) run is selected (orange) and can be used for loss landscape experiments. The vertical axis represents the number of epochs in thousands (depending on training settings).}
    \label{fig:lossepoch}
\end{figure}

Each experiment starts with setting up and training the network (T1-Finding interesting points). The first step is to create a random target point in the sampling menu (\autoref{fig:teaser} (i)) within the selected parameter range as an initial state. The new target point is then visible in the target point menu (\autoref{fig:teaser} (h)), together with the loss value and L2 norm (weight vectors with very small or large magnitude could be difficult to train), and can now be selected. The user starts training from the selected target point by clicking on the "Start training" button. After training, the newly created target points are shown in the loss-epoch diagram, see \autoref{fig:lossepoch}. It is also possible to create multiple runs, in which case each one will be assigned a new color in the line chart and the target point table.

\subsection{Global landscape} \label{fig:globallandscape}

\begin{figure}[htp]
    \centering
    \subfigure[]{\includegraphics[width=0.16\textwidth]{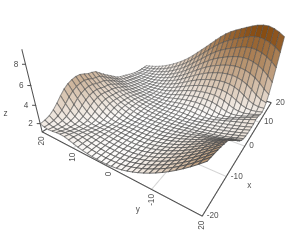}} 
    \subfigure[]{\includegraphics[width=0.16\textwidth]{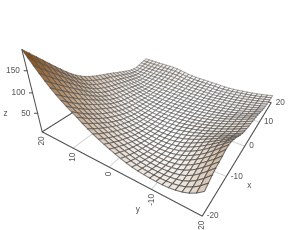}}
    \subfigure[]{\includegraphics[width=0.16\textwidth]{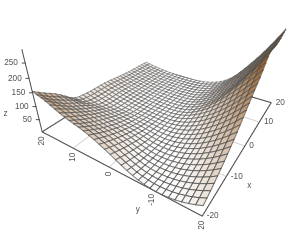}}\\
    \subfigure[]{\includegraphics[width=0.16\textwidth]{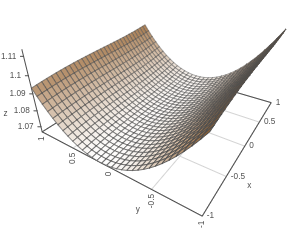}}
    \subfigure[]{\includegraphics[width=0.16\textwidth]{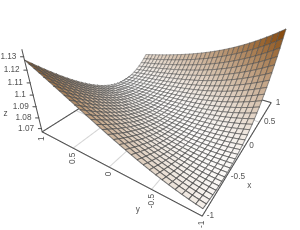}}
    \subfigure[]{\includegraphics[width=0.16\textwidth]{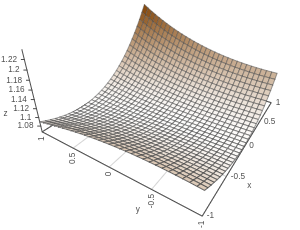}}
    \caption{Six 2D slice plots show that the area around the 0-vector (at the origin) is flat, regardless of the scale or axis vectors of the plane. The random vector approach generates a new plot each time, which is why multiple samples are given. (a), (b), and (c) cover a large area (side length of 40), (d), (e), and (f) cover a small area (side length of 2) around the origin. Note that all visualizations indicate a flat region at the center, and the loss value changes minimally for (d), (e), (f).}
    \label{fig:surfaceplots}
\end{figure}

We start with the question whether we could visualize the general landscape within a given parameter range, similarly to \autoref{fig:himmelblau} and \autoref{fig:ackley}.
We plot the loss around the zero vector (every weight parameter is 0).
The surface plots in \autoref{fig:surfaceplots} are created by selecting the zero vector target point and clicking on "Plot (random directions)" (\autoref{fig:teaser} (d)) and show 2D slices of the loss surface around the 0-vector. We can see that there are directions where the loss value increases more strongly than in others, but the area at the origin appears to be within a flat region. We also notice that sampling a larger space around the origin creates a more turbulent landscape (more small minima along the axes) than sampling a smaller range. The marginal loss value changes in the areas depicted in \autoref{fig:surfaceplots} (d), (e), and (f) confirm that the region around the loss is relatively flat.

\begin{figure}[htp]
    \centering
    \subfigure[Slices in the loss range from 0 to 50.]{\includegraphics[width=0.49\textwidth]{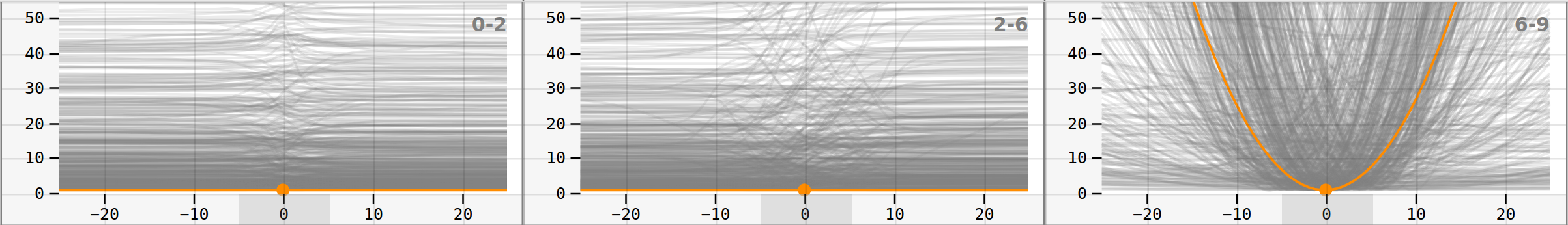}} \\
    \subfigure[Slices in the loss range from 0 to 2.]{\includegraphics[width=0.49\textwidth]{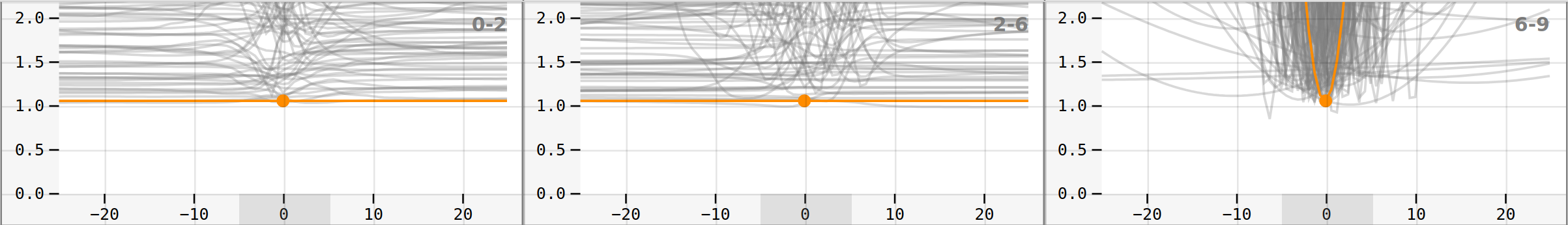}}
    \caption{Slice charts for the first weight parameter in each layer (from neuron 0 to neuron 2, 2 to 6, and 6 to 9 as shown in \autoref{fig:smallnetwork}). Grey slices are created from focus points. The orange slices are created from the target point (the zero vector). An increased slice density towards lower loss values can be observed, but only a few sampled slices reach a loss value below 1.0.}
    \label{fig:zeroslices}
\end{figure}

Next, we use the Sobol algorithm to create 500 focus points in the range from -5 to 5 and slice each parameter from -25 to +25 with a resolution of 81 samples/slice. We show three representative slice charts in \autoref{fig:zeroslices} (since the remaining 28 appear redundant in this particular case). We make the following observations: (1) All slice charts within a network layer appear to be qualitatively similar, but not between layers. This can be explained by the fact that all weights at the zero vector have the same value, which generally leads to similar behavior between parameters.
(2) The slice charts appear approximately symmetric, with some of the slices ascending and others descending. We assume that this is caused by other parameters becoming negative or positive, which affects the influence of this weight.
(3) At some distance from 0, the parameters in the first two layers show no change regardless of whether they are made smaller or larger. This can be explained by the Sigmoid activation function, which limits the output of each neuron between 0 and 1.
(4) Few slices reach a loss below 1.0, and none of the slices reaches a loss close to 0. This indicates that the area has such a small density of low loss regions, compared to high loss regions (loss larger than 1), that random sampling and slicing do not find any weight vector which produces a low loss.

While \autoref{fig:surfaceplots} perturbs parameters randomly, and each weight is taken into account to a different degree, \autoref{fig:zeroslices} reveals all components that are implicitly contained in \autoref{fig:surfaceplots} without emphasizing some weights over others. In particular, \autoref{fig:zeroslices} shows that the weights in the last layer, which uses a linear activation function, add a parabolic component to the loss landscape, while the first two layers appear to add comparably flat regions.

\subsection{Minima} \label{ch:minima}

\begin{figure}[htp]
    \centering
    \subfigure[Loss 0.16]{\includegraphics[width=0.16\textwidth]{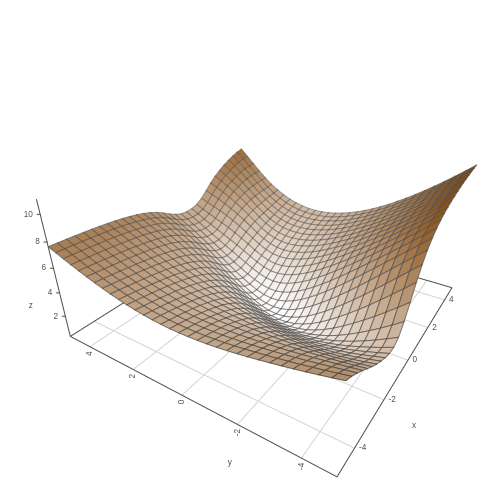}}
    \subfigure[Loss 0.07]{\includegraphics[width=0.16\textwidth]{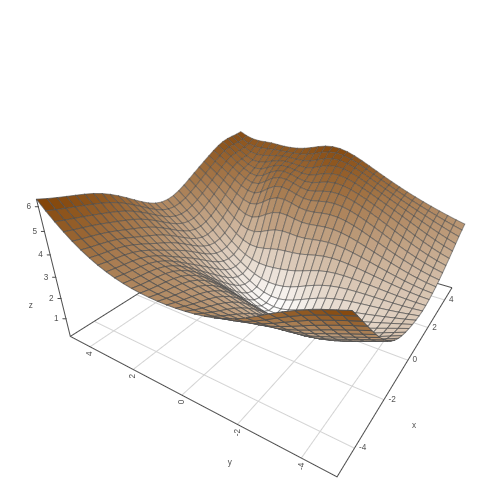}}
    \subfigure[Loss 0.03]{\includegraphics[width=0.16\textwidth]{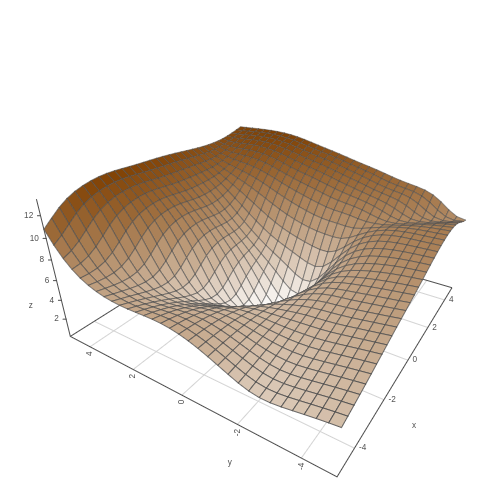}}\\
    \subfigure[Loss 1.0]{\includegraphics[width=0.1\textwidth]{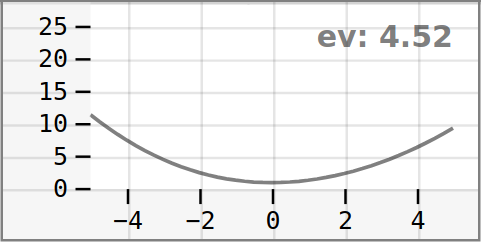}}
    \subfigure[Loss 0.23]{\includegraphics[width=0.1\textwidth]{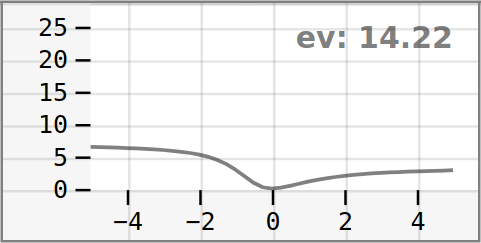}}
    \subfigure[Loss 0.08]{\includegraphics[width=0.1\textwidth]{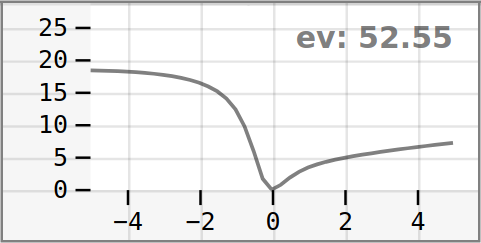}}
    \subfigure[Loss 0.02]{\includegraphics[width=0.1\textwidth]{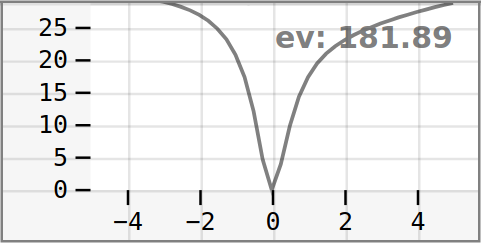}}
    \caption{The funnel-like shape becomes more significant as the network converges to a minimizer. In the first row, we see the minimizers for different loss values on random 2D planes. In the bottom row, we see the minimizers that become visible with our 1D EV slicing approach. The slices follow the direction of the most significant EV.}
    \label{fig:surfaceplots2}
\end{figure}

\begin{figure}[htp]
    \centering
    \subfigure[Weight slices]{\includegraphics[width=0.38\textwidth]{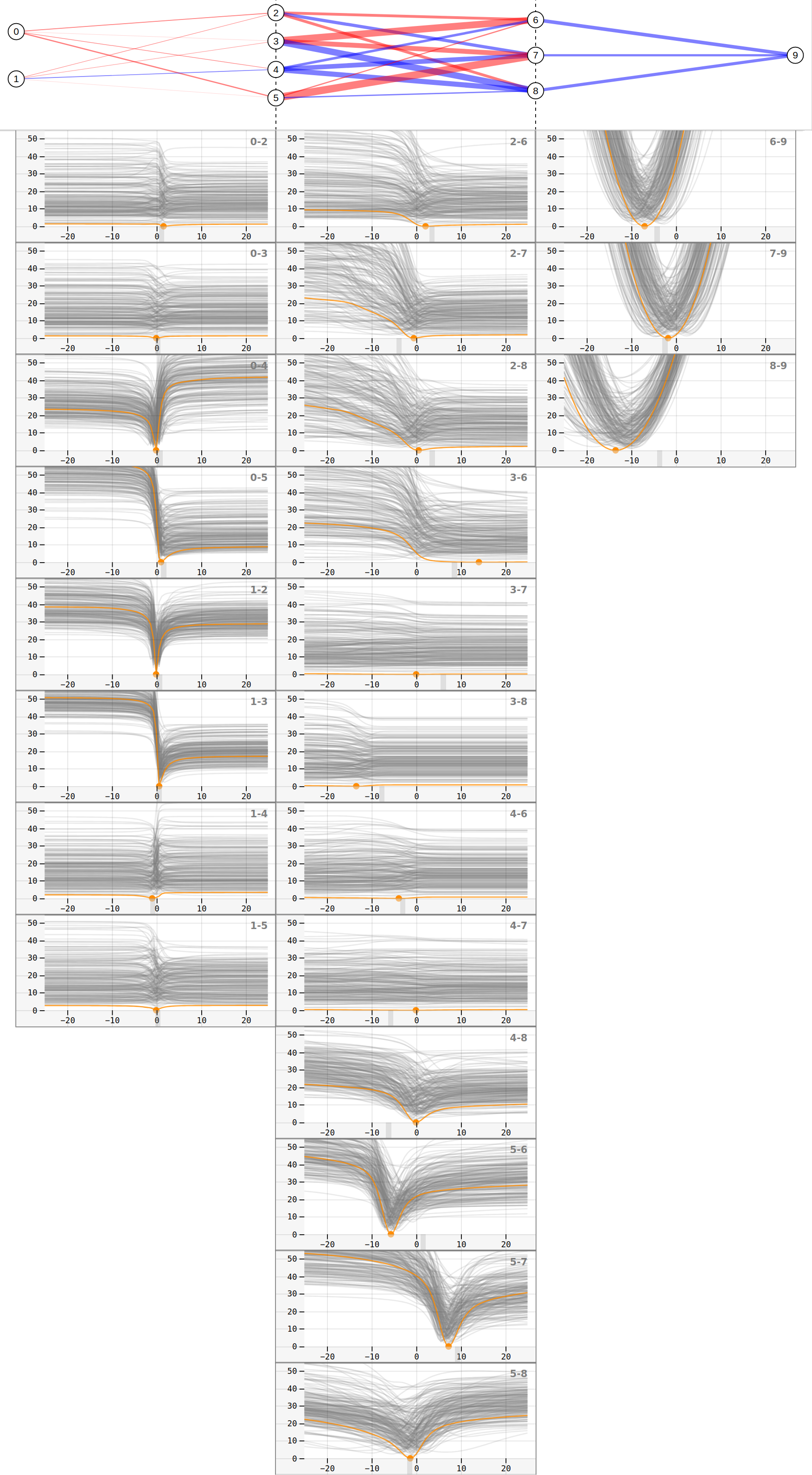}}
    \subfigure[Bias slices]{\includegraphics[width=0.10\textwidth]{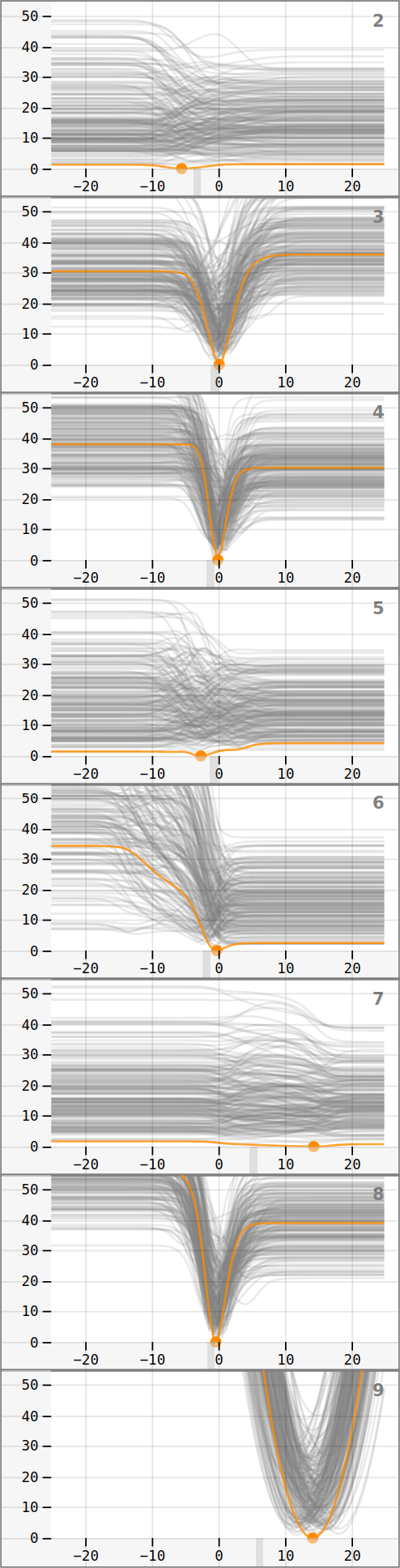}}
    \caption{A minimizer with a loss of 0.04 is visualized by 31 slice charts, one for each network parameter. The slice charts correspond to the weights in the model above. Unlike the slicing around the zero vector (\autoref{fig:zeroslices}), the slices around the minimum reveal different surface shapes for different weight dimensions even within the same layers.}
    \label{fig:fullsliecchart}
\end{figure}

\begin{figure}[htp]
    \centering
    \subfigure[Ground truth data]{\includegraphics[width=0.18\textwidth]{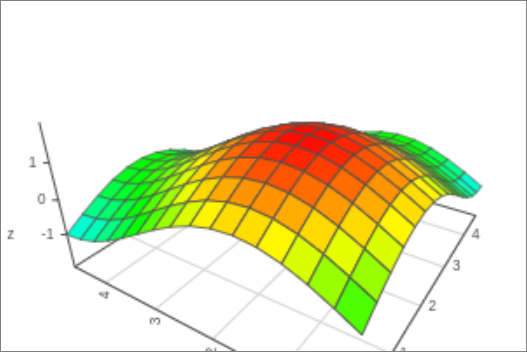}}\\
    \subfigure[Prediction with loss 0.36]{\includegraphics[width=0.13\textwidth]{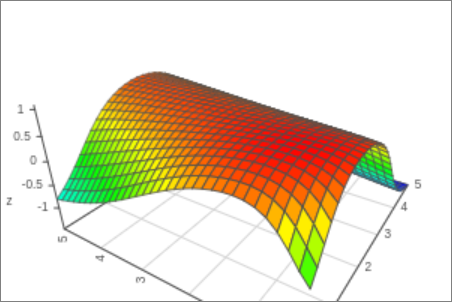}}
    \subfigure[Prediction with loss 0.18]{\includegraphics[width=0.13\textwidth]{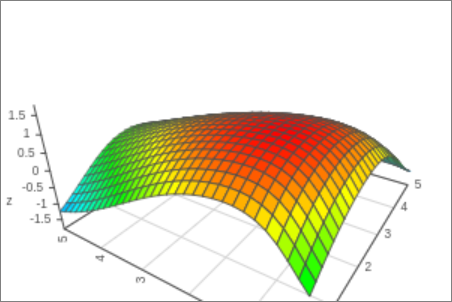}}
    \subfigure[Prediction with loss 0.04]{\includegraphics[width=0.13\textwidth]{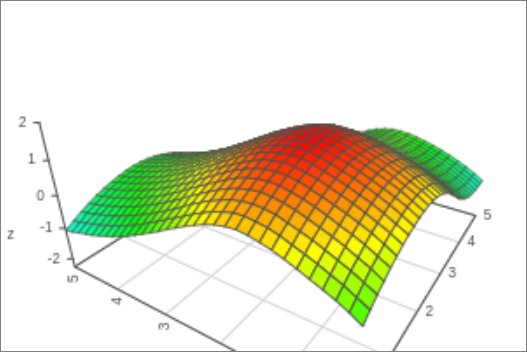}}
    \caption{Network predictions can be plotted for each target point. Note that height and color both encode the output value (heatmap).}
    \label{fig:predictions}
\end{figure}

\begin{figure}[htp]
    \centering
    \includegraphics[width=0.24\textwidth]{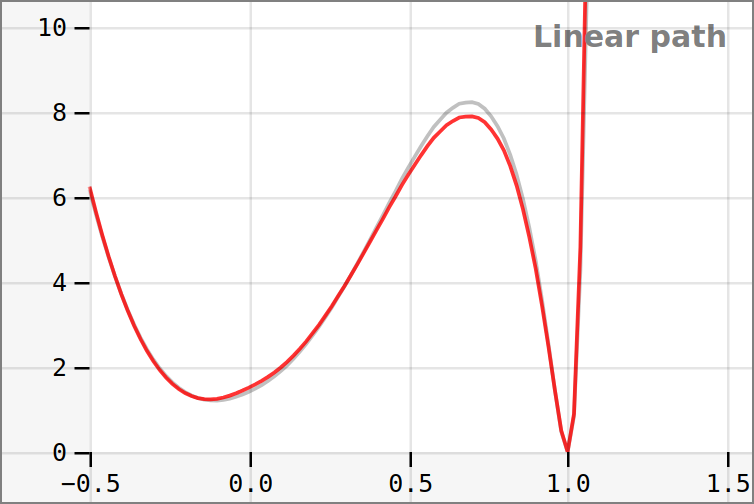}
    \caption{The Adam algorithm avoided a high-loss barrier during training from loss value 1.42 to 0.04. The loss values are not in the visualization, but they are visible in the menu where the selection of the two target points was made (\autoref{fig:teaser} (h)). The local minimum is behind the starting parameter (=0) and has no particular meaning in this case.}
    \label{fig:linearpaths}
\end{figure}

The familiar shapes of minima that are common in 1D and 2D functions can also be reproduced in high-dimensional spaces. \autoref{fig:surfaceplots2} shows how different slicing methods change the representation of minimizers. We observe that the funnel shape becomes more significant during training. However, the slice charts in \autoref{fig:fullsliecchart} show a more detailed view of the loss landscape and reveal function characteristics that have been concealed by \autoref{fig:surfaceplots2}. We can see that, for example, weights 0-4, 0-5, 1-2, and 1-3 are more sensitive to small changes than weights 3-8 and 4-6, for this particular minimizer, which is not visible in the visualization from \autoref{fig:surfaceplots2}. We also observe that, for example, weights 3-7 and 4-7 are less sensitive to changes than other weights, but if we zoom in, we can see that they still contribute a small part to the prediction. Another observation is that the loss landscape in the first two layers is full of apparently flat areas, which could be an obstacle for gradient-based algorithms, especially second-order algorithms.

There are two additional ways to observe target points: \autoref{fig:predictions} shows the network output for different training stages. This output is used to compute the loss value. Unfortunately, this approach is not possible for networks with a greater number of input and output neurons, but it helps the user understand the meaning of the loss value for this small network model. It is also possible to plot the linear interpolation path from a high loss to a lower loss value, as in \autoref{fig:linearpaths}. We observe a wide high-loss minimizer on the interpolation axis, as well as a relatively sharp minimizer at 1.0.







\subsection{Batch gradient descent}

\begin{figure}[tp]
    \centering
    \subfigure[Weight slices]{\includegraphics[width=0.38\textwidth]{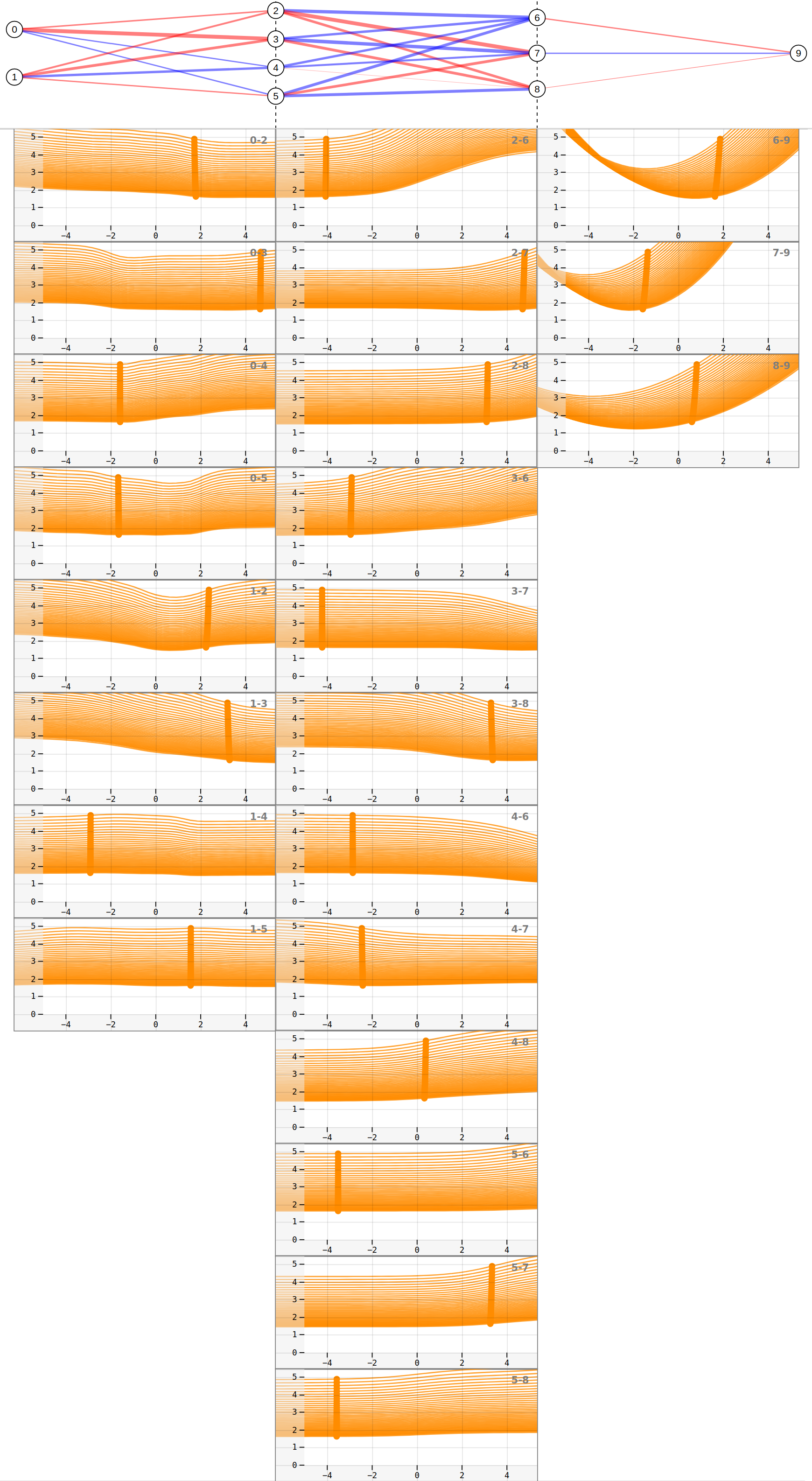}}
    \subfigure[Bias slices]{\includegraphics[width=0.10\textwidth]{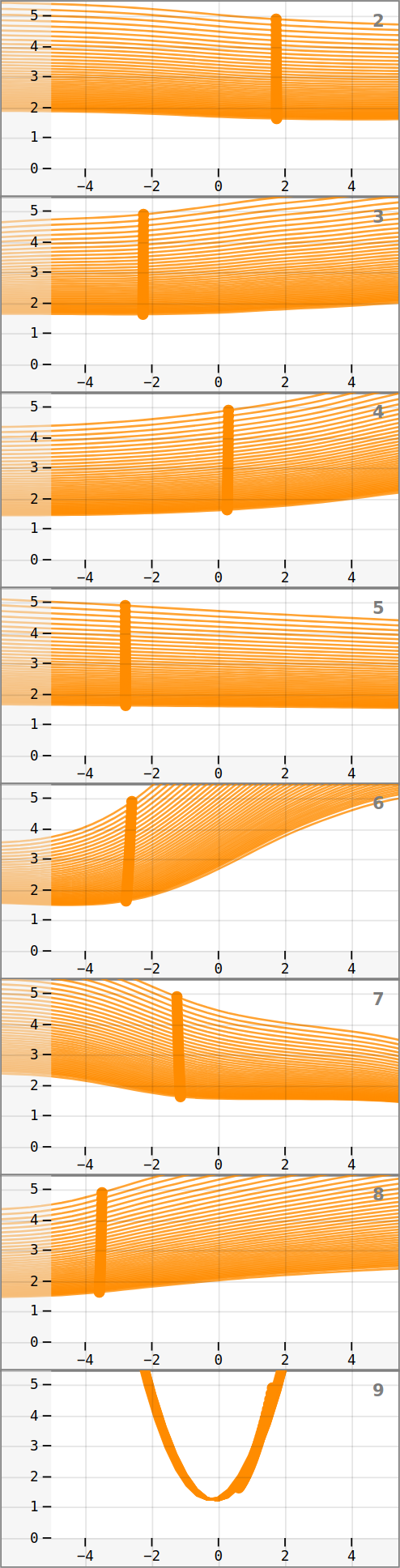}} \\
    \subfigure[Zoomed bias slices for neuron 9]{\includegraphics[width=0.24\textwidth]{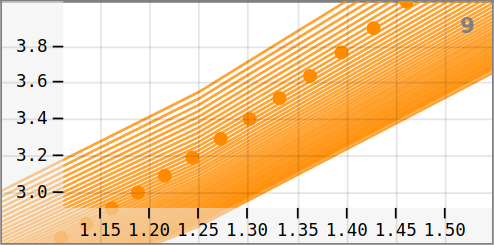}}
    \caption{Early training stages: Each target point and the respective slices correspond to one of the first 50 training epochs of gradient descent training with a learning rate of 0.01. During these 50 epochs, the loss value decreases from 4.89 to 1.63 along a relatively smooth and simple loss surface. Unlike the previous slice charts, these visualizations only show target points (which are always highlighted in orange) and their axis-parallel slices instead of focus points (which produce grey slices).}
    \label{fig:sgdweightsfull}
\end{figure}

\begin{figure}[htp]
    \centering
    \subfigure[Linear splines]{\includegraphics[width=0.24\textwidth]{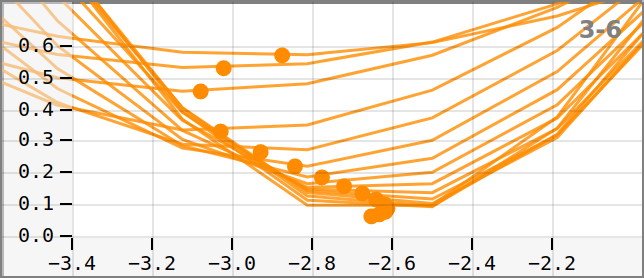}}
    \subfigure[Natural splines]{\includegraphics[width=0.24\textwidth]{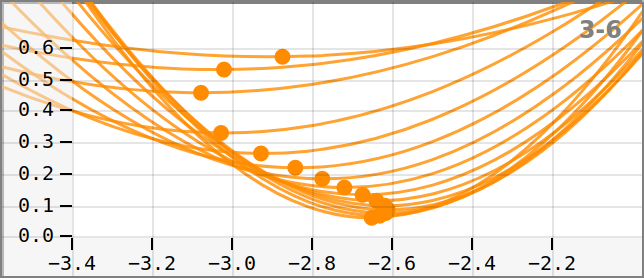}}
    \caption{The slice chart of weight 3-6 for target points between 0.57 and 0.06 shows that gradient descent with lr=0.01 avoids areas of high loss while following a downward trajectory. The natural splines fill missing data points to make the visualization more intuitive.}
    \label{fig:traj36}
\end{figure}

In this use case, we apply the slicing technique to visualize the training trajectory of batch gradient descent by slicing the target points along the training trajectory. The first target point, which is used as the initial weight vector, is sampled from a uniform distribution in the range [-5,5]. 
We use gradient descent with a learning rate (lr) of 0.01 and train on the full data. \autoref{fig:sgdweightsfull} shows a slice chart visualization for the first 50 training epochs where the loss is reduced from 4.89 to 1.63. We observe that the loss is significantly reduced even though the weights only change a little. A large change, compared to the other parameters, can be seen in the output layer bias (\autoref{fig:sgdweightsfull} (b) and (c)). Overall we observe a low curvature of the surface but also a steep descent which gradually slows down towards lower loss values, where the gaps between the slices become smaller. The low curvature could partly be caused by the last bias parameter approaching its own minimum and lowering the loss for all other slices.

In \autoref{fig:sgdweightsfullfinal} we observe a later stage in training where the loss is reduced from 0.57 to 0.06. The slices are now $1,000$ epochs apart because the convergence slows down towards the end. We make the following observations:

\begin{figure}[tp]
    \centering
    \subfigure[Weight slices]{\includegraphics[width=0.38\textwidth]{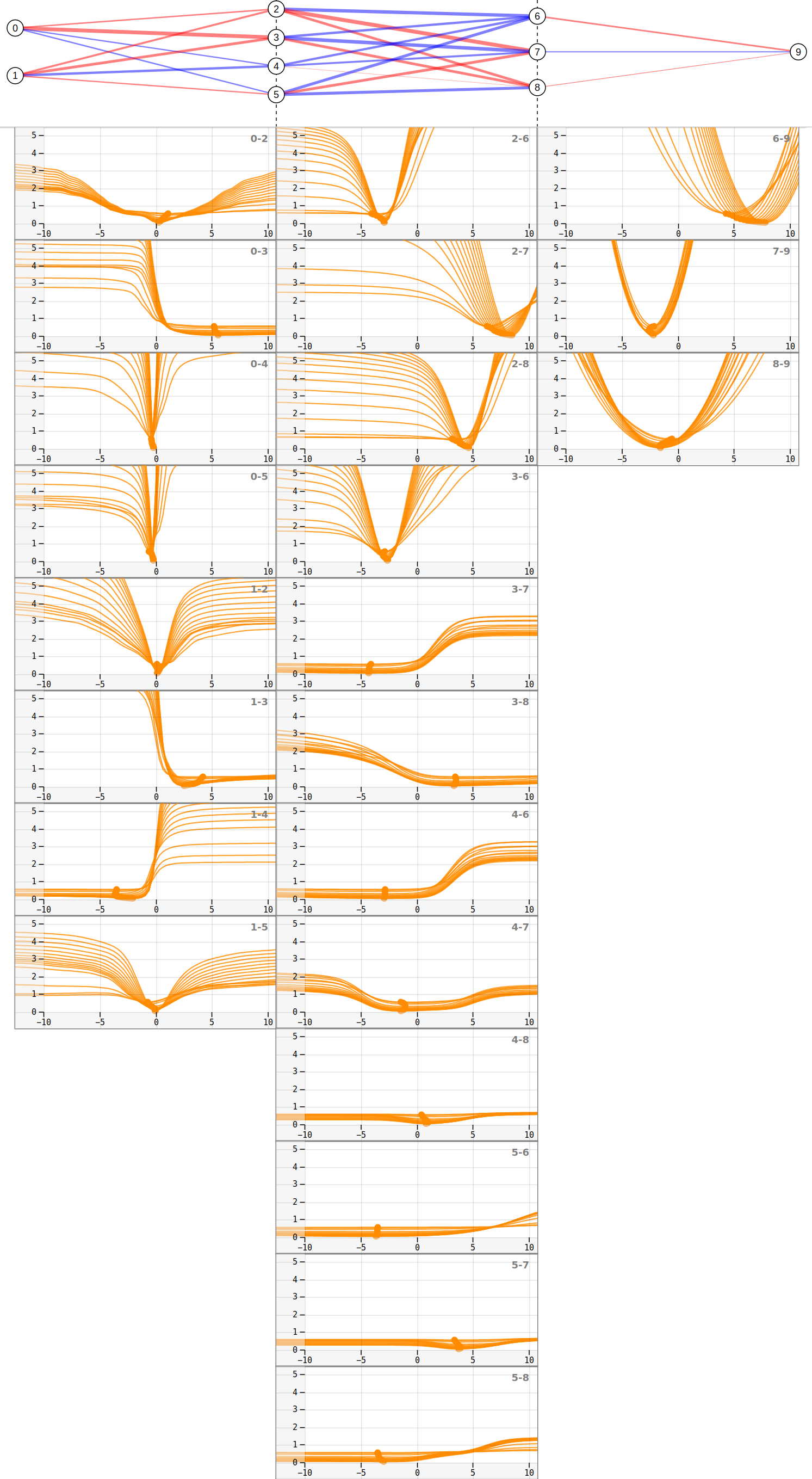}}
    \subfigure[Bias slices]{\includegraphics[width=0.10\textwidth]{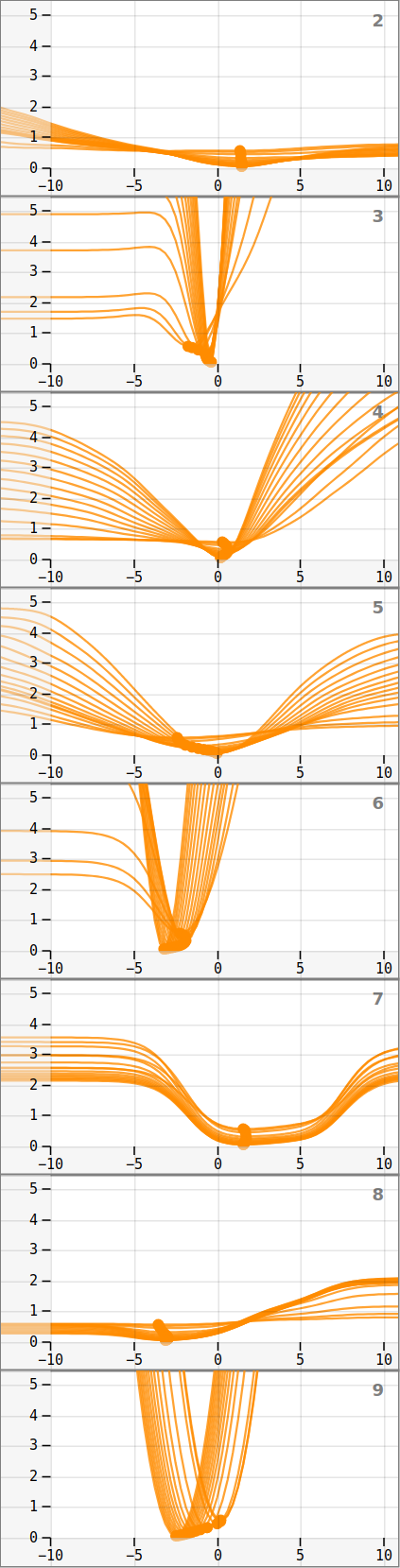}}
    \caption{Late training stages: Gradient descent training (learning rate 0.01) with the loss value decreasing from 0.57 to 0.06. The slices are more diverse than at the beginning of the training (\autoref{fig:sgdweightsfull}). All target points appear to be at the minimum of their respective axis-parallel slices. A zoomed-in version of slice chart 3-6 is shown in \autoref{fig:traj36}.}
    \label{fig:sgdweightsfullfinal}
\end{figure}


\paragraph{The landscape geometry becomes more complex} The slices have stronger curvatures and are more complex towards the end of the training (in \autoref{fig:sgdweightsfullfinal}) than they were at the beginning of the training (in \autoref{fig:sgdweightsfull}). This appears to be related to the observation from \autoref{ch:minima}, which shows a stronger curvature as the algorithm approaches a minimizer. 

\paragraph{Gradient descent appears to navigate obstacles} In \autoref{fig:sgdweightsfullfinal} we see that the optimization algorithm moves along valleys (since all target points remain at the lowest part of the axis slices throughout the training). This aligns with expectations of the gradient descent algorithm and validates the visualization approach. However, the observation of the descent from all directions in the high-dimensional space on their actual axes is a novelty. \autoref{fig:traj36} shows in detail how a weight parameter adapts to changes by following the descent direction along a curved valley, avoiding high-loss regions. As the training progresses, the remaining descent direction lies off-axis. Otherwise, they would be visible in the slice charts. Instead, we see that the algorithm is always at the lowest point in the slice. Note that this is not the case at the beginning of the training (\autoref{fig:sgdweightsfull}), where we can still see lower parameter values along the axes. High-loss barriers that appear on linear interpolations, such as seen in \autoref{fig:linearpaths}, could be avoided in a similar way.
\section{User Evaluation}  \label{ch:evaluation}

In addition to our case studies in the previous section, we also gathered feedback from external experts. We presented the final prototype to four ML practitioners: one physicist and three computer scientists. After providing a tutorial PDF and letting them try out the tool, we asked what they liked/did not like, whether they believed the prototype could be useful, and which features they would add. There was an agreement that the sampling menu, as well as the slice charts, were found overwhelming or hard to interpret (4/4). However, most users also stated that the prototype could definitely be useful (3/4) in different fields, in particular, for autoencoders, ensemble networks, and theoretical research. Slicing the Hessian EV directions was considered an interesting and novel visualization method (2/4). Both researchers who found the EV plots interesting stated that they frequently computed Hessian EVs but had no method to analyze them in detail. Other user comments were that the integration of different visualization methods in a single dashboard makes it easier to interpret the visualizations in a greater context (1/4), that the visualizations of the target function and the model predictions are easy to understand (1/4), and that exploring different points of the training process in detail appears useful (1/4). Requested features were tools to analyze the slices in more detail, to add axis labeling, and an improved loss-epoch chart.

After trying out the prototype, the test users answered the System Usability Score \cite{brooke1996sus} questionnaire, which resulted in scores of 65.0, 67.5, 70.0, and 75.0, respectively. The mean SUS score is close to the average score of 68, which is neither good nor bad. From a usability point of view, there is room for improvement, especially in the context of making the slice visualizations more interpretable and user-friendly.


\section{Discussion}  \label{ch:discussion}

Understanding the loss landscape is essential for understanding the convergence behavior of DNNs and, hence, for the understanding of network architectures, setup, and initialization. It has been extensively studied in the ML community such that whole workshops are dedicated to loss landscapes \cite{epfl}, yet the vis community has ignored it so far. This is in spite of the fact that, at its core, it is a challenge in understanding high-dimensional scalar functions. Hence, one of our major contributions is to introduce the vis community to this problem. Further, our requirement analysis has been done through the study of the previous (ML) literature, since it is a rich literature, clearly articulating the need and research questions. Further, the problem is extremely complex, such that there should be no expectation that a single paper will solve all issues. Hence, this paper is making a first attempt at looking at alternative approaches to visualize this high-dimensional function, known as the loss landscape. Therefore, we decided to build a prototype that explores, based on small networks, whether the principle idea of axis-aligned slices is a feasible approach to understand the loss landscape.

In that regard, we showed that the slicing approach can be applied to the 31D loss surface of a DNN, that visualizations with meaningful axes can be created, and that patterns can be found within such visualizations.
We succeeded in developing a prototype that allows the users to follow a fast trial-and-error approach of setting up experiments and explore the loss landscape from different perspectives and in great depth that, as far as we know, has not been reached by any previous tools or literature.

\paragraph{What are the advantages of the slice charts?}
We learned that one advantage of slice charts over random 2D slice plots is that a view can be reproduced by sampling focus points over the same region again at some point in the future. It is therefore not necessary to save the random direction vectors or to recompute the random vector visualizations until a good view is found. Additionally, every point in the loss landscape can be investigated from the same perspectives, while random direction plots appear different each time they are computed for the same weight vector. Slice charts are also robust, i.e., they show if the general geometry of the network stays the same, even for different minimizers where weights have swapped their meaning due to symmetries in the loss landscape. Additionally, they show visualizations for different layers, which can reveal differences in loss landscape geometry between layers. Another advantage is that slice charts are an analysis method for DNNs that only requires forward propagation and a loss function to be computed, i.e., no modifications of the model are needed. This makes them suitable for black box models in general, where derivatives cannot be computed but where the optimization landscape can still be sampled. 
In other words, slice charts give a truly global view of the high-dimensional landscape. The downside of this approach is the relatively large number of single charts one has to manage. Therefore, at this stage, it is a valuable tool for training environments (similar to the Tensorflow Playground \cite{playground}), to get an intuition of how different aspects of a network behave. Scaling this approach to large, real-world DNNs remains a challenge and is left for future research.

Hence, our original question (\emph{Are axis-aligned slices a feasible approach to understand high-dimensional loss landscapes?}) can be answered in the affirmative. However, there are a number of limitations that have to be overcome in order to apply this to real-world DNNs.


\paragraph{Would this approach work for large networks?}
The presented approach is intractable for networks with millions of parameters, but a small subset of network units could be selected or systematically sampled. Selecting weights based on heuristics could be possible. The analysis could also start with a few parameters and low-resolution slices and let the user add more weights or slices on demand. Finally, based on the use case in \autoref{fig:globallandscape} and the qualitative similarities of the slices per layer, we can form the hypothesis that, in the case of the zero vector, it would be sufficient to plot a single slice chart per layer, which would make this approach applicable for larger network architectures with dozens of layers. Sophisticated analysis tools for the slice charts would be required to capture patterns and untangle the slices.

\section{Conclusion}  \label{ch:conclusion}

Loss landscape experiments help researchers understand DNNs, but visualization of high-dimensional surfaces is a challenge and often requires reducing dimensionality and creating axes without obvious meaning. Linear interpolation experiments, for example, show barriers that are not actually encountered during training \cite{DBLP:conf/icml/DraxlerVSH18}, and random 2D slice plots show convexity \cite{DBLP:conf/nips/Li0TSG18} but not geometry of the loss surface. Inspired by Sliceplorer \cite{DBLP:journals/cgf/Torsney-WeirSM17} and Tensorflow Playground \cite{playground}, we developed a prototype that allows the user to set up experiments and analyze the loss landscape from different perspectives. We demonstrated how slice charts show the loss surface in greater detail than 2D slices around a minimizer and the zero vector, we showed how to apply the slicing approach to EV directions, and we showed how batch gradient descent traverses the landscape in the early stages of training and follows descent directions through valleys in late stages of training. The user evaluation indicates that further research is needed to improve the usability and interpretability, but that our approach could be useful for analyzing autoencoders and ensemble networks.

\subsection{Future work}
The greatest trade-off for the slicing approach is the number and complexity of the visualizations. However, one has to keep in mind that there are no good alternatives for high-dimensional scalar functions. For large networks, slice charts could only be computed for selected parts of the network. Further, to make slice charts more readable, it could help to cluster (groups of) slices. Hessian EV analysis using the slicing approach could be explored further, especially in the context of second-order optimization.

We are also in the process of making our web-based tool available for free online to gather more feedback from the ML community.


\bibliographystyle{abbrv-doi}

\bibliography{template}
\end{document}